\documentclass[11pt]{amsart}

\usepackage[letterpaper, margin=1in]{geometry}

\usepackage{amsmath, amssymb, amsthm, amsaddr}
\usepackage{enumerate, subcaption, graphicx, hyperref}

\title{Time-varying Transition Matrices with Multi-task Gaussian Processes}
\author{Ekin Ugurel} 

\address{University of Washington, Seattle, WA 
\\ \texttt{ugurel@uw.edu}}

\date{\today} 

\DeclareMathOperator*{\argmin}{arg\,min}
\begin{document}

\maketitle
\begin{abstract}
In this paper, we present a kernel-based, multi-task Gaussian Process (GP) model for approximating the underlying function of an individual's mobility state using a time-inhomogeneous Markov Process with two states: moves and pauses. Our approach accounts for the correlations between the transition probabilities by creating a covariance matrix over the tasks. We also introduce time-variability by assuming that an individual’s transition probabilities vary over time in response to exogenous variables. We enforce the stochasticity and non-negativity constraints of probabilities in a Markov process through the incorporation of a set of constraint points in the GP. We also discuss opportunities to speed up GP estimation and inference in this context by exploiting Toeplitz and Kronecker product structures. Our numerical demonstrate the ability of our formulation to enforce the desired constraints while learning the functional form of transition probabilities. 
\end{abstract}

\section{Introduction}
\label{introduction}
Understanding human mobility has long been a subject of interest in various academic disciplines, tracing its roots back to the early works of geographers and anthropologists \cite{binford1990mobility, macdonald1999reproductive}. Over the years, researchers have employed various methodologies, ranging from surveys and interviews to mathematical models, to unravel the complexities of human movement patterns. However, with the advent of the digital age and the widespread use of smartphones, a wealth of passively-generated mobile data has become available. This abundance of data has revolutionized the field of human mobility modeling, enabling researchers to delve deeper into the intricacies of human behavior and uncover previously hidden insights. Consequently, numerous studies have emerged, leveraging these datasets to explore diverse aspects of human mobility, such as transportation planning \cite{poonawala2016singapore}, urban commute patterns \cite{frias2012estimation}, and epidemic spread \cite{hao2020understanding}, among others.

One prominent approach used in modeling human mobility is the application of Markov processes, with Hidden Markov Models (HMMs) being particularly relevant in this field. HMMs have been widely employed to capture the sequential nature of human movement and understand the underlying patterns \cite{mathew2012predicting, shi2019semantics}. By considering a sequence of observed locations as a series of states, HMMs facilitate inference of the hidden states or latent variables that govern human mobility. This methodology has proven valuable in predicting future locations, identifying distinct behavioral patterns, and detecting anomalies in mobility patterns. While traditional HMMs assume stationarity, recent research has explored the use of time-inhomogeneous Markov processes to account for temporal variations in human mobility \cite{jiang2016timegeo}. These time-inhomogeneous models take into consideration the evolving dynamics of movement patterns over time, allowing for more accurate and flexible representations of human mobility behaviors. 

This paper presents a novel approach to address the challenge of accurately estimating dynamic transition probabilities in human mobility modeling using passively-generated mobile data, with a specific focus on the temporal dimension. To overcome the issue of irregularly sampled data, the proposed method leverages multi-task Gaussian Processes (GPs) to learn and predict correlated transition probabilities as functions of time. Furthermore, to enforce the stochasticity and non-negativity constraints inherent in Markov processes, our approach uses a set of constraint points that act as reference points where the constraints are enforced, allowing us to enforce these constraints functionally within the GP framework. This approach, explored previously in \cite{pilar2022incorporating, pensoneault2020nonnegativity}, provides a robust and accurate estimation of dynamic transition probabilities, enhancing our understanding of human mobility patterns and enabling more reliable predictions.

\section{Dataset and Algorithm Implementation}\label{sec:algorithms}
We use a passively-generated mobile dataset from \href{https://spectus.ai}{Spectus} to derive dynamic transition matrices as functions of time. This dataset contains pointwise observations of various individuals in the Greater Seattle Area over a six-month period in 2020. The raw data includes user IDs, timestamps, latitudes and longitudes, and a measure of precision (i.e., a spatial radius for which the provider has 95\% confidence in the reported coordinates). 

We implemented our methodology in the \textsc{Python} programming language \cite{CS-R9526}. We used \textsc{GPyTorch} \cite{gardner2018gpytorch} to increase efficiency and speed up matrix inversions within our GP framework. Furthermore, we used \textsc{pandas} \cite{mckinney-proc-scipy-2010} to read and wrangle the mobile dataset from which we derived the time-dependent transition matrices and \textsc{NumPy} \cite{harris2020array} for a variety of array operations. Last but not least, we used \textsc{matplotlib} 
 \cite{Hunter:2007} to produce any visualizations of our results.

\section{Model Formulation}
In this section, we describe the stochastic process model for human mobility in the context of a Gaussian Process. The latter part of the section describes opportunities to exploit structure while using temporal input dimensions in order to speed up the inversion of the covariance matrix
\subsection{Two-state Markov Process of Human Mobility}
\label{methodology}
We model the change in one's mobility state using a time-inhomogeneous Markov Process over a general state space with two states: moves and pauses. Let $P_m(t)$ be the probability that an individual will be in the move state at time $t$, and let $a_{pm}(t)$ be the probability of switching from the pause state at time $t-1$ to the move state at time $t$.  Then, the probabilities $P_m$ are related over time by
$$
    P_m(t) = a_{pm}(t)P_p(t-1) + a_{mm}(t)P_m(t-1)
$$
or in matrix notation
$$
    P(t) = a(t)P(t-1)
$$
where $P(t)$ is a 2-dimensional vector of state probabilities and $a(t)$ is a $2\times2$ matrix of transition probabilities, $a_{ij}(t)$.  Since the two states are mutually exclusive, the elements of $P(t)$ and the rows of $a(t)$ must sum to $1$ in every period (this is the stochasticity constraint):
\begin{equation}
\label{StochasticityConstraint}
    \begin{aligned}
        a_{pp}(t) + a_{pm}(t) &= 1 \\
        a_{mp}(t) + a_{mm}(t) &= 1 \\
        P_{p}(t) + P_{m}(t) &= 1
    \end{aligned}
\end{equation}

We can now introduce time-variability. We assume that an individual's transition probabilities vary over time in response to exogenous variables, such as time or location. That is,
\begin{equation}
\label{eq:time-variability}
    \begin{aligned}
        a_{pm}(t) &= f_{pm}(z(t-1), \beta_{pm}) \\
        a_{pp}(t) &= f_{pp}(z(t-1), \beta_{pp}) \\
        a_{mp}(t) &= f_{mp}(z(t-1), \beta_{mp}) \\
        a_{mm}(t) &= f_{mm}(z(t-1), \beta_{mm}) \\
    \end{aligned}
\end{equation}
where $z(t-1)$ is a vector of exogenous variables at time $t-1$ and $B_{ij}$ for $i, j \in \{m, p\}$ is a vector of parameters relating the variables to the transition probabilities. Since the transition probabilities must be non-negative and satisfy the stochasticity constraint (Eq. \ref{StochasticityConstraint}), the choice of functional forms for $f_{ij}$ in Eq. \ref{eq:time-variability} can be limited.  For example, one common formulation is the multinomial logit. We, however, do not make such assumptions but rather rely on a non-parametric formulation based on Gaussian Processes.

\subsection{Multi-task Gaussian Process}

Let $t_i, m_i$ be discrete timestamps and observed mobility states for $i = 1, \ldots, N$. We aggregate observations at each hour of the day and each day of the week to derive transition matrices specific to those temporal intervals. For example, assume that we observe ten instances of a person's mobility habit from 9AM to 10AM on Monday (i.e., 10 weeks). In this period, the person embarks on a trip while previously being paused 6 weeks out of 10, and finishes a trip within the same period (i.e., transitions from moving to pausing) 4 times out of 10. For this individual, $a_{pp} = 0.4, a_{pm} = 0.6, a_{mp} = 0.4, a_{mm} = 0.6$, and so on. 

We store the transition probabilities in a matrix $$\mathbf{A} = [\mathbf{a_{pp}}, \mathbf{a_{pm}}, \mathbf{a_{mm}}, \mathbf{a_{mp}}]$$ where each row of the matrix corresponds to a time interval. We note that since we only have two states, our supervised learning problem boils down to learning one entry from each row, as the latter can be estimated by subtracting the former from $1$. Assume the underlying data generating process for each transition probability is $\mathbf{a} = \mathbf{f(t)} + \epsilon$ where $\mathbf{f} = [f_1, \ldots, f_N]^T$ specifies a systematic function of timestamps $t_i$ and $\epsilon = [\epsilon_1, \ldots, \epsilon_N]^T$ is a white noise process. Both $\mathbf{f}$ and $\epsilon$ are assumed to be normally distributed
\begin{equation*}
    \begin{aligned}
        \mathbf{f} &\sim \mathbb{N}(\mathbf{f} | \mu(t), \mathbf{K}) \\
        \epsilon &\sim \mathbb{N}(\epsilon | 0, \sigma_a^2 \mathbf{I})
    \end{aligned}
\end{equation*}
where $\mu(t)$ is the constant mean function (i.e., taken as the average of the training set), $\mathbf{K} = k(\mathbf{t}, \mathbf{t})$ is a positive semi-definite (PSD) kernel (or covariance matrix), and $\sigma_y^2$ is the variance of the noise. We can then write the likelihood as
\begin{equation*}
    \mathbf{A} | \mathbf{t}, \sigma_a^2 \sim \mathbb{N}(\mathbf{a}| \mu(t), \mathbf{K} + \sigma_a^2\mathbf{I})
\end{equation*}
An inferred probability $a_*$ of a new timestamp $t_*$ conditioned on the training data also has a Gaussian distribution and can be computed in closed form as
\begin{equation*}
    a_* | t_*, \mathbf{t}, \mathbf{A}, \sigma_a^2 \sim \mathbb{N}(a_* | \mu_*, \sigma_*^2)
\end{equation*}
where
\begin{equation}
\label{regularinference}
    \begin{aligned}
        \mu_* &= k(t_*, \mathbf{t})(\mathbf{K} + \sigma_a^2 \mathbf{I})^{-1}\mathbf{A}, \\ 
        \sigma_*^2 &= k(t_*, t_*) - k(t_*, \mathbf{t})(\mathbf{K} + \sigma_a^2 \mathbf{I})^{-1} k(\mathbf{t}, t_*)
    \end{aligned}
\end{equation}
Furthermore, this formulation has a unique particularity where the four dependent variables are highly correlated due to the stochasticity constraint. Therefore, instead of treating as independent outcomes, we can account for the correlations by creating a covariance matrix over the transition probabilities \cite{bonilla2007multi}. This covariance structure for the four tasks (i.e., the output vector with four scalars) can be specified as
\begin{equation*}
    \mathbf{K}(\mathbf{a}_{ij}, \mathbf{a}_{ij}) = k(t_*, \mathbf{t}) \mathbf{K}^f(\mathbf{a}_{ij}, \mathbf{a}_{ij})
\end{equation*}
where $\mathbf{a}_{ij}$ for $i \in \{m, p\}$ is one of four transition probability vectors, $\mathbf{K}^f$ is a PDS matrix containing the inter-task covariance and $k$ is any valid PDS kernel. A key property of this model is that the joint normal distribution over the entire output space is not block-diagonal with respect to tasks. That is, observations of one task can affect the predictions on another task. To account for the newly-introduced $\mathbf{K}^f$ term in our inference, we update Equation \ref{regularinference} as follows
\begin{equation*}
        \begin{aligned}
        \pmb{\mu_*} &= (k_l^f \otimes k_*)(\mathbf{K}^f \otimes \mathbf{K} + D \otimes \mathbf{I})^{-1}\mathbf{A}, \\ 
        \pmb{\sigma_*^2} &= (k_l^f \otimes k_{**}) - (k_l^f \otimes k_*)(\mathbf{K}^f \otimes \mathbf{K} + D \otimes \mathbf{I})^{-1} (k_l^f \otimes k_*)
    \end{aligned}
\end{equation*}
where $\otimes$ denotes the Kronecker product, $k_l^f$ selects the $l^{th}$ column of $\mathbf{K}^f$, $k_* = k(t_*, \mathbf{t})$ is the vector of covariances between the test point and the training set, $D$ is an $M \times M$ diagonal matrix in which the $(l, l)^{th}$ element is the noise variance for the $l^{th}$ task (i.e., $\sigma_l^2$), and $k_{**} = k(t_*, t_*)$. 
Finally, rather than using simple cross-validation, we choose to minimize the negative marginal log-likelihood in determining the optimal hyperparameters $\Theta$.
\begin{equation}
\begin{aligned}
\label{eq:mll}
    -\log(p(\mathbf{A}|\mathbf{t}, \Theta)) &= \frac{1}{2}[\mathbf{A}^T (\mathbf{K} + \sigma_a^2)^{-1} \mathbf{A} + \log|\mathbf{K}| + N\log(2\pi)]
\end{aligned}
\end{equation}
where $\Theta$ is the set of model parameters (i.e., kernel parameters like the lengthscale). Keeping in mind the stochasticity constraint as well as the non-negativity constraint of probabilities, Eq. \ref{eq:mll} fits into the following constrained optimization problem.
\begin{equation}
\begin{aligned}
\label{eq:unconstrained GP}
\argmin_{\Theta} \quad & -\log(p(\mathbf{A} | \mathbf{t}, \Theta)) \\
\textbf{s.t.} \quad & a_{pm}^*(t) + a_{pp}^*(t) = 1 \hspace{0.2in}\forall t \in N\\
\quad & a_{mp}^*(t) + a_{mm}^*(t) = 1 \hspace{0.2in}\forall t \in N \\
\quad & a_{ij}^*(t) \geq 0 \hspace{0.2in}\forall i, j \in \{m, p\}
\end{aligned}
\end{equation}
In a Markov Process with two states, the stochasticity constraint can be enforced rather straightforwardly (i.e., we can just estimate one of the probabilities and subtract it from one to determine the other), but we cannot say the same about the non-negativity constraint---since it is a functional constraint, it can be difficult to enforce in practice. To get around this, we enforce the constraints in Eq. \ref{eq:unconstrained GP} on a set of constraint points $\mathbf{t}_c = \{t_c^{(u)}\}_{u=1}^m$. Subsequently, to obtain the constrained GP, we solve the following constrained minimization problem
\begin{equation}
\begin{aligned}
\label{eq:constrained GP}
\argmin_{\Theta} \quad & -\log(p(\mathbf{A} | \mathbf{t}, \Theta)) \\
\textbf{s.t.} \quad & a_{pm}^*(t_c^{(u)}) + a_{pp}^*(t_c^{(u)}) = 1 \hspace{0.21in} \forall u = 1, \ldots, m\\
\quad & a_{mp}^*(t_c^{(u)}) + a_{mm}^*(t_c^{(u)}) = 1 \quad \forall u = 1, \ldots, m\\
\quad & a_{ij}^*(t_c^{(u)}) \geq 0 \quad \hspace{0.15in}\forall i, j \in \{m, p\} \hspace{0.1in} \textrm{and} \hspace{0.1in} u = 1, \ldots, m
\end{aligned}
\end{equation}

\subsection{Toeplitz Structure Exploitation using Kronecker Product}
Approximating a function to fit a dynamic transition matrix is generally not an expensive procedure, but there may be cases in which additional input/output dimensions are used and training time is increased dramatically. To overcome this, the structure of the training data can be exploited to speed up matrix operations. 

A time-inhomogeneous Markov Process lies on an equally-spaced (perfectly observed, albeit noisy) temporal grid. In our case, we group observations by the day of the week and hour of the day---hence, we choose to represent time in a 2D grid. $\mathbf{K}$ is a Toeplitz covariance matrix if it is generated from a stationary covariance kernel, $k(\mathbf{t}, \mathbf{t'}) = k(\mathbf{t} - \mathbf{t'})$, with inputs $x$ on a regularly spaced one dimensional grid. Toeplitz matrices are constant along their diagonals: $\mathbf{K}_{i,j} = \mathbf{K}_{i+1,j+1} = k(\mathbf{t}_i - \mathbf{t}_j)$. In higher dimensions, a multi-dimensional grid can be expressed as the Kronecker product of Toeplitz matrices \cite{gardner2018product}.

The speed-up of this formulation comes from certain linear algebra properties. First, Toeplitz matrices can be embedded as circulant matrices to perform fast matrix vector products using fast Fourier transforms, reducing the inversion operation complexity from $\mathcal{O}(n^3)$ to $\mathcal{O}(n\log n)$. Second, the full covariance matrix can be expressed as the Kronecker product of two Toeplitz matrices (one for each grid dimension) due to the following linear algebra property:
$$(\mathbf{A} \otimes \mathbf{B})^{-1} = \mathbf{A}^{-1} \otimes \mathbf{B}^{-1} $$

Thus, for a two-dimensional grid of size $nm \times nm$, the complexity of estimating the inverse of the diagonal matrix is $\mathcal{O}(n\log n + m\log m)$, rather than the usual $\mathcal{O}(n^3m^3)$.

\section{Numerical Results}
\label{numericalresults}

\begin{figure}
     \centering
     \begin{subfigure}[b]{1.0\textwidth}
         \centering
         \includegraphics[width=\textwidth]{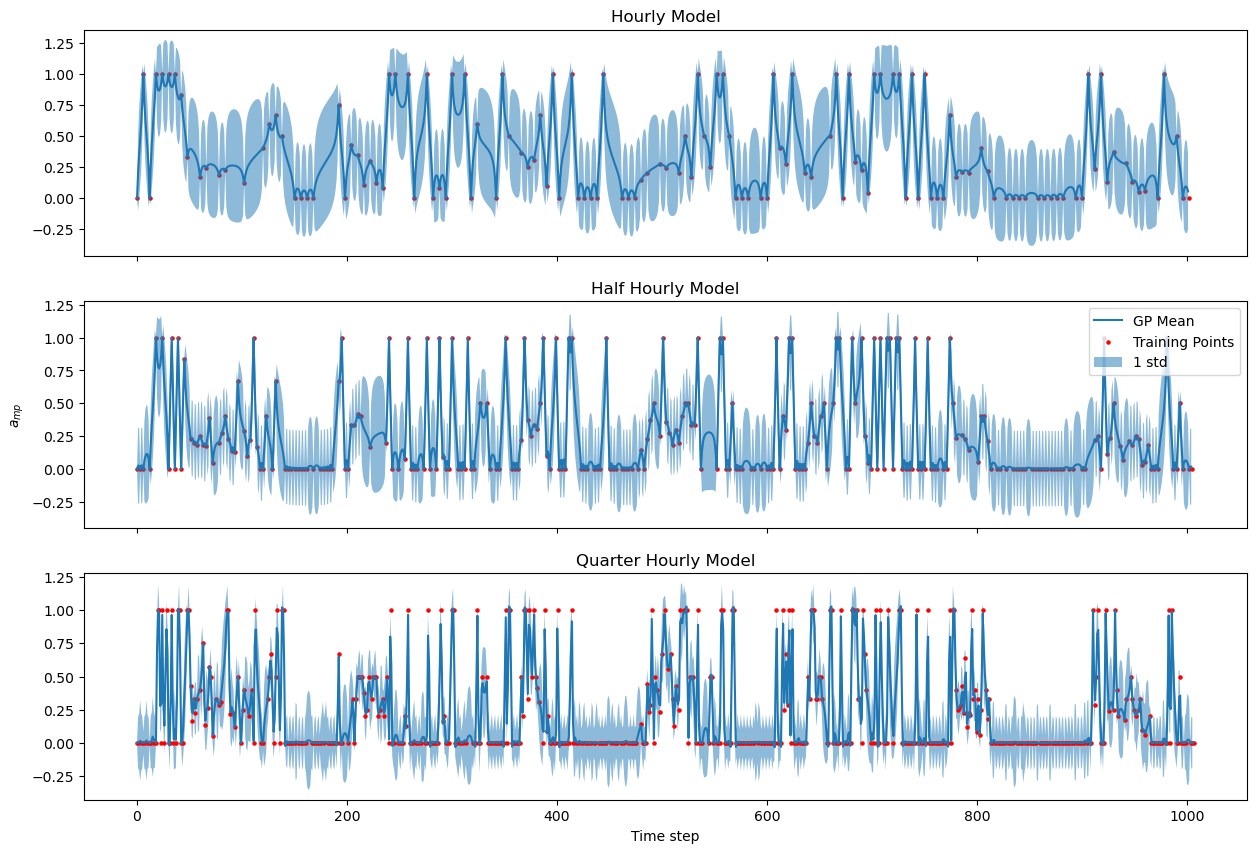}
         \caption{Noise-free modeling of $a_{mp}$}
         \label{fig:noise-free}
     \end{subfigure}
     \begin{subfigure}[b]{1.0\textwidth}
         \centering
         \includegraphics[width=\textwidth]{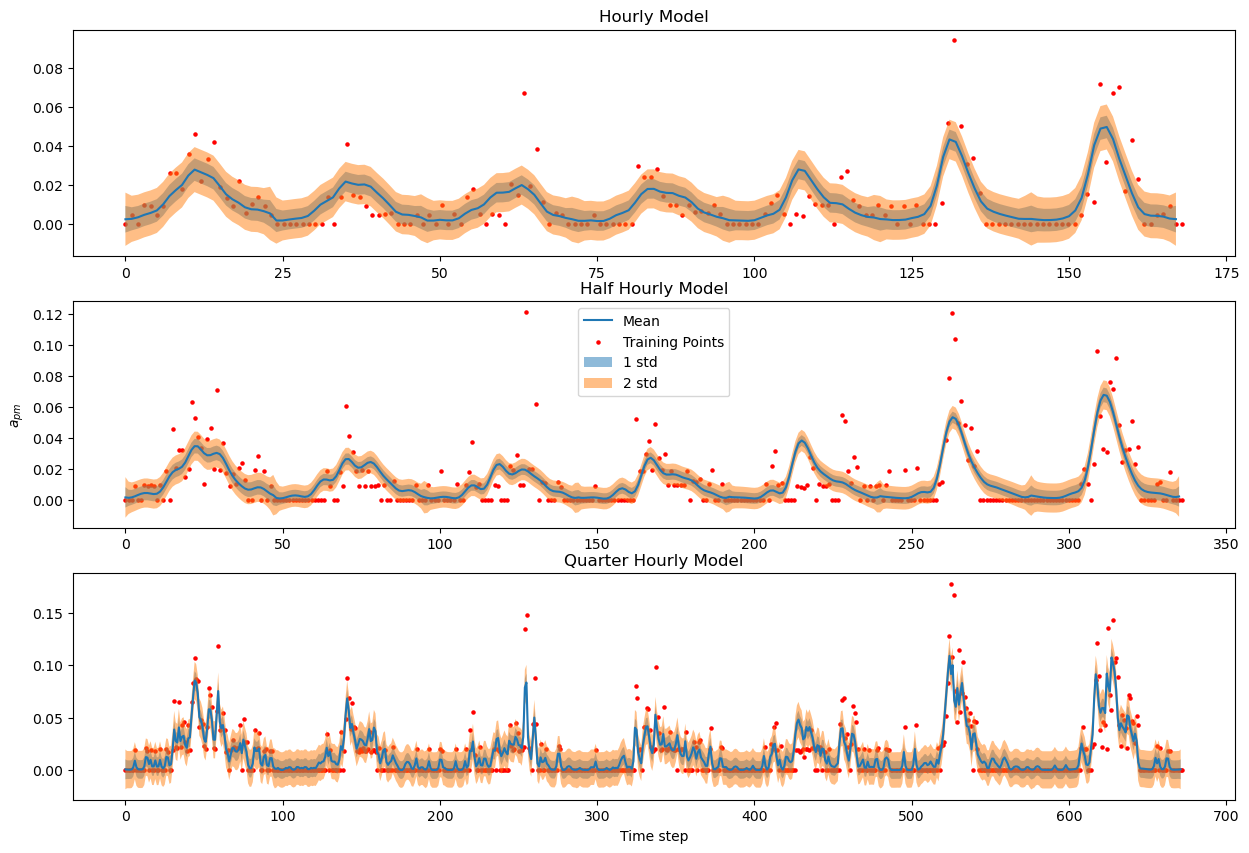}
         \caption{Grid-based modeling of $a_{pm}$ with noise}
         \label{fig:gridKernelwithnoise}
     \end{subfigure}
        \caption{}
        \label{fig:1}
\end{figure}

We solve the minimization problem shown in Eq. \ref{eq:constrained GP}. In our experiments, we tried different variations of the core model involving differences in how we specify noise and representing temporal dimensions in a grid. These are shown in Figure \ref{fig:1}: \ref{fig:noise-free} shows a noise-free\footnote{In our implementation, noise is not equal to zero, but rather a very small number. Zero noise leads to numerical instabilities in the inversion of the covariance matrix.} model of one of the transition probabilities, while \ref{fig:gridKernelwithnoise} shows a noisy (i.e., Gaussian noise) model using a grid structure for temporal dimensions. Here, we notice a few things:
\begin{itemize}
    \item A noise-free GP model resembles a simple curve-fitting model with linear polynomials
    \item A GP with Gaussian noise is less prone to outliers in the training data
    \item As we go from an hourly model to a quarter-hourly model, the variance of the GP becomes more constrained, a further proof that we are effectively enforcing the non-negativity and stochasticity constraints of the Markov Process
\end{itemize}

Informal experiments using grid-based time representation exploiting Toeplitz structure showed a range of training time reductions from 45\% to 66\%. In the end, however, we decided not to include grid-based models into our framework due to the difficulty of comparing their test accuracy against non-grid-based models. Because our formulation only allows estimating transition matrices at discrete time intervals, it is not possible to exclude any testing points in fitting a grid-based GP model (otherwise, it wouldn't be a proper grid!).

\begin{figure}
     \centering
     \begin{subfigure}[b]{0.85\textwidth}
         \centering
         \includegraphics[width=\textwidth]{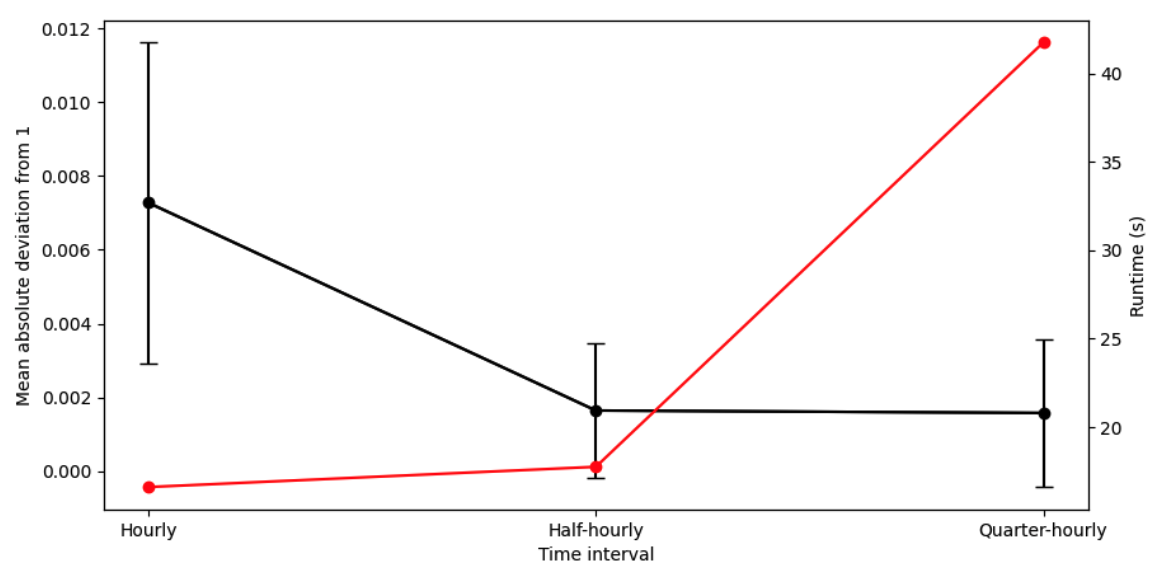}
         \caption{Stochasticity constraint satisfaction (left axis) and runtime (right axis) per level of transition probability discretization}
         \label{fig: constraintSat}
     \end{subfigure}
     \begin{subfigure}[b]{0.6\textwidth}
         \centering
         \includegraphics[width=\textwidth]{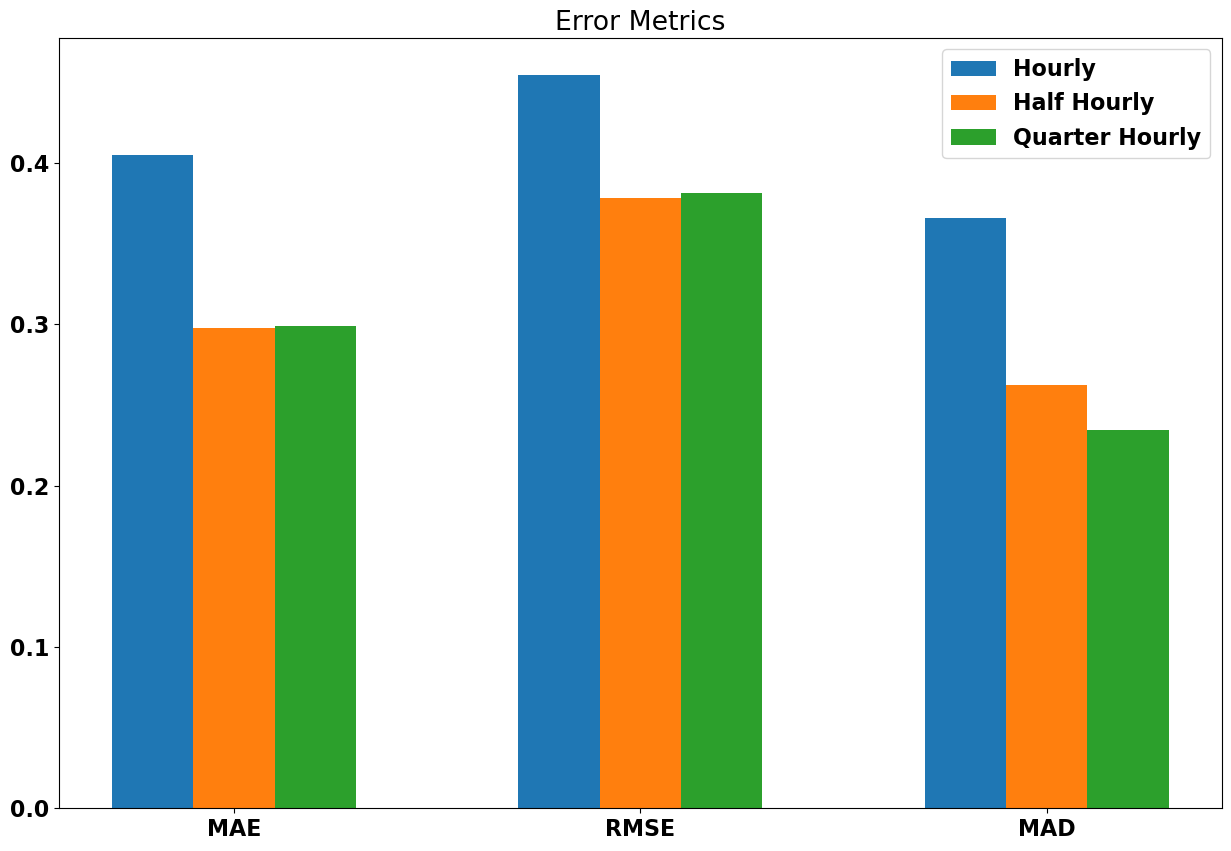}
         \caption{Error metrics per level of transition probability discretization}
         \label{fig: errors}
    \end{subfigure}
    \begin{subfigure}[b]{0.8\textwidth}
        \centering
        \includegraphics[width=\textwidth]{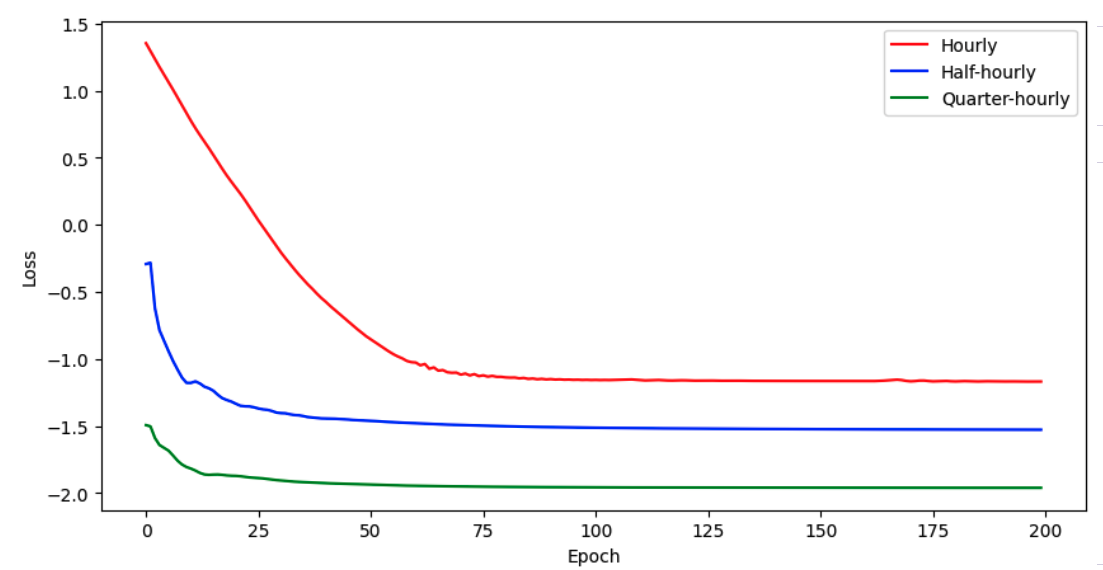}
        \caption{Loss function progression per level of transition probability discretization}
        \label{fig: lossProg}
    \end{subfigure}
        \caption{}
        \label{fig:2}
\end{figure}

Figure \ref{fig:2} shows the results of introducing additional constraint points in the GP. Specifically, Figure \ref{fig: constraintSat} suggests that as we go from hourly to quarter-hourly transition matrices, we achieve a greater satisfaction of the stochasticity constraint while also introducing more computational complexity (notice the cubically increasing runtime as the number of training points double). 

Figure \ref{fig: errors} shows the testing accuracy of various discretization levels. Noticeably, we do not observe significant improvements in accuracy going from half-hourly  to quarter-hourly intervals. This indicates that there may be a tradeoff between increasing the number of time intervals estimated in a time-inhomogeneous Markov process and model generalizability---that is, we hypothesize a theoretical upper bound on accuracy improvements able to be achieved from decreased discretization of a stochastic process.

Finally, Figure \ref{fig: lossProg} shows the progression of the loss function for the three levels of discretization. The quarter-hourly model achieves the lowest loss and also converges much quicker than the hourly model, while the half-hourly model is roughly in betweenn the two. A more careful mathematical analysis here could potentially prove convergence rates in the context of Markov processes.

\newpage
\section{Summary and Conclusions}\label{sec:conclusions}
We have presented a novel method for approximating the underlying function of a time-inhomogeneous Markov process using multi-task Gaussian Processes. We have enforced the stochasticity and non-negativity constraints of this Markov process using a set of constraint points on the GP, essentially discretizing the process into smaller chunks. We have discovered that this approach comes with certain tradeoffs, including added costs in computational complexity. We have also hypothesized the existence of an upper bound on the accuracy benefits to be gained from decreased discretization, which may add more noise to the Markov process. 

Future research efforts on this topic could explore alternative methods to enforce the stochasticity and non-negativity constraints of a stochastic process. These may include alterations to the loss function of a GP to penalize the model for predictions that do not satisfy the desired constraints. Additionally, methods to speed up GP estimation and inference in the context of time-inhomogeneous stochastic processes should be given careful consideration. Finally, the aforementioned tradeoffs between discretization and noise deserve a closer look, as well as convergence rates of different time-inhomogeneous Markov processes (these two points are related). 

\section*{Acknowledgements}
The authors are thankful to Prof. Hosseini for useful discussions about the project scope. The authors also note that only Ekin Ugurel had access to the Spectus dataset, from which the time-dependent transition matrices were generated.

\newpage
\bibliographystyle{abbrv}
\bibliography{references}

\end{document}